\documentclass[journal]{IEEEtran}
\IEEEoverridecommandlockouts
\usepackage{amsmath,amssymb,amsfonts}
\usepackage{algorithmic}
\usepackage{graphicx, hyperref, url}
\usepackage{textcomp}
\usepackage{csquotes}
\usepackage{xcolor,soul}
\usepackage{adjustbox}
\usepackage{hyperref}
\usepackage{multirow}
\usepackage{titlesec}
\usepackage{subcaption}
\usepackage{makecell}
\usepackage{amsfonts}

\usepackage[numbers,sort&compress]{natbib}

\usepackage{etoolbox}

\definecolor{columbiablue}{rgb}{0.61, 0.87, 1.0}
\titlespacing{\subsubsection}{0pt}{0pt}{1pt}

\usepackage{amsmath}

\usepackage{mathtools}
\usepackage[colorinlistoftodos,prependcaption,textsize=tiny]{todonotes}
\usepackage{gensymb}
\usepackage{multirow}
\usepackage{orcidlink}
\usepackage{parskip}

\begin{document}
\title{Unscented Kalman Filter with a Nonlinear Propagation Model for Navigation Applications}
\author{Amit Levy \orcidlink{0009-0009-8703-5807},
Itzik~Klein \orcidlink{0000-0001-7846-0654}
\thanks{A. Levy and I. Klein are with the Hatter Department of Marine Technologies, Charney School of Marine Sciences, University of Haifa, Israel.\\ Corresponding author: alevy02@campus.haifa.ac.il}
}

\maketitle

\begin{abstract}
The unscented Kalman filter is a nonlinear estimation algorithm commonly used in navigation applications. The prediction of the mean and covariance matrix is crucial to the stable behavior of the filter. This prediction is done by propagating the sigma points according to the dynamic model at hand. In this paper, we introduce an innovative method to propagate the sigma points according to the nonlinear dynamic model of the navigation error state vector. This improves the filter accuracy and navigation performance. We demonstrate the benefits of our proposed approach using real sensor data recorded by an autonomous underwater vehicle during several scenarios.
\end{abstract}

\begin{IEEEkeywords}
Inertial Navigation, Kalman filter, Nonlinear filters, sensor fusion 
\end{IEEEkeywords}

\section{Introduction}\label{intro_sec}
The linear Kalman filter is an optimal minimum mean squared error estimator for state estimation of linear dynamic systems in the presence of Gaussian distribution and Gaussian noises. For nonlinear problems, the extended Kalman filter (EKF) is used. The EKF linearizes the system and measurement models using a first-order partial derivative (Jacobian) matrix while applying the KF propagation step (or the update step) to the error-state covariance matrix \cite{barShalomappandtrack2004}. The linearization procedure introduces errors in the posterior mean and covariance, leading to sub-optimal performance and at times divergence of the EKF \cite{WanUkfnonlin2000}. To circumvent the problem, Julier and Uhlmann \cite{JUlUKF1997} introduced the unscented Kalman filter (UKF). The underlying idea was that a probability distribution is relatively easier to approximate than an arbitrary nonlinear function. To this end, UKF uses carefully chosen, sigma points that capture the state mean and covariance. When propagated through the nonlinear system, it also captures the posterior mean and covariance accurately to the third order \cite{WanUkfnonlin2000}. \\
Generally, the UKF is more accurate than the EKF and obviates the need to calculate the Jacobian matrix. Like the EKF, it requires the prior statistical characteristic of system noise, process noise covariance, and the measurement noise covariance to be precisely known, specifically because the covariances directly regulate the effect of prediction values and measurements on system state estimations. Describing the exact noise covariance, which may change during operation, is a challenging task but an important one because uncertainty may reduce the filter estimation performance or even lead to its divergence. Therefore, methods for adjusting the process and measurement noise covariance matrices have been suggested in the literature, introducing the concept of adaptive filters.\\
Adaptive EKF algorithms were designed to learn the process noise covariance online \cite{kleinhybridekf2022, Nadavkleinhybridekf2024, EKFCM2012, EKFAUVCM2016}. Recently, an adaptive neural UKF (ANUKF) has been introduced, capable of tuning the process noise matrix dynamically and thereby coping with the nonlinear characteristics of the inertial fusion problem \cite{ANUKF2025}.\\
Regardless if the filter is adaptive or not, the UKF uses a dynamic model to propagate the state vector and the covariance matrix. Having an inaccurate dynamic model can decrease performance and may lead to filter divergence. When applying the UKF to improve the performance of a navigation system, some authors used the navigation solution as the state vector as in \cite{AUVACOUSTIC2018} and \cite{FederatedUkf2021}. It is also common to use the navigation error solution as the state vector, enhancing the accuracy of the unscented transform \cite{AUVHypersonic2020}. There, a model predictive UKF was introduced for inertial navigation system/global navigation satellite system (INS/GNSS) fusion.\\
In this paper, we introduce the unscented Kalman filter navigation error state propagating method (UKF-NESPM) to accurately propagate the sigma points using the navigation algorithm itself, enabling us to capture the exact dynamic functionalities. We demonstrate the benefits of our proposed approach using real sensor data recorded by an autonomous underwater vehicle (AUV) during several scenarios.\\
The rest of the paper is organized as follows: Section \ref{prob_form_sec} describes the UKF algorithm.
Section \ref{prop_appr} introduces our proposed method.
Section \ref{res_sec} presents our AUV dataset and results. Finally, Section \ref{conc_sec} provides the conclusions of this work.

\section{Problem Formulation}\label{prob_form_sec}
\subsection{The UKF algorithm}\label{ukf_std_alg}
Consider UKF to estimate the unobserved state vector {x} of the following dynamic system:
\begin{equation}\label{general_dynamic_model_eq}
    \centering
    \boldsymbol{x}_{k+1} = {f}(\boldsymbol{x}_{k}, \boldsymbol{\omega}_{k})
\end{equation}
where ${x}_{k}$ is the unobserved state at step ${k}$, ${f}$ is the dynamic model mapping, and ${\omega}$ is the process noise.
\begin{equation}\label{general_meas_model_eq}
    \centering
    \boldsymbol{z}_{k+1} = {h}(\boldsymbol{x}_{k+1}, \boldsymbol{\nu}_{k+1})
\end{equation}
where, ${h}$ is the observation model mapping, ${\nu}$ is the measurement noise, and ${z_{k+1}}$ is the observed signal at step ${k+1}$.
UKF \cite{JUlUKF1997} is based on the scaled unscented transform (UT), a method for calculating the statistics of a random variable that undergoes a nonlinear transformation. It starts with a selected population, propagates the population through the nonlinear function, and computes the appropriate probability distribution (mean and covariance) of the new population. The population is carefully selected to induct the mean and covariance of the undergoing population by the nonlinear transformation.
The UKF is an iterative algorithm. The initialization step is followed by iterative cycles, each consisting of several steps. Step 1 is the initialization, steps 2-4 are executed iteratively (step 4 upon measurement availability). These steps are described below.

\begin{enumerate}
    \item Initialization\\
        Given the initializing random ${n}$ state vector ${x}_{0}$ with known mean and covariance, set
        \begin{equation}\label{init_x_eq}
            \centering
            \hat{\boldsymbol{x}}_{0} = {E[{x}_{0}]}
        \end{equation}
        \begin{equation}\label{init_p_eq}
            \centering
            \boldsymbol{\mathbf{P}}_{0} = {E[ ({x}_{0}-\hat{\boldsymbol{x}}_{0})({x}_{0}-\hat{\boldsymbol{x}}_{0})^{T} ]}
        \end{equation}
        set the weights for the mean (uppercase letters m) computation and the covariance (uppercase letters c):
         \begin{equation}\label{w_m_0_eq}
            \centering
            \mathbf{w}_{0}^{m} = \frac{\lambda}{(n+\lambda)}
        \end{equation}
         \begin{equation}\label{w_c_0_eq}
            \centering
            \mathbf{w}_{0}^{c} = \frac{\lambda}{(n+\lambda)} + {(1 + \alpha^2 + \beta)}
        \end{equation}
         \begin{equation}\label{w_i_eq}
            \centering
            \mathbf{w}_{i}^{m} = \mathbf{w}_{i}^{c} = \frac{1}{2(n+\lambda)}, {i=1,...,2n}
        \end{equation}
            
        \item Update sigma points\\
        Set ${2n+1}$ sigma points as follows:
         \begin{equation}\label{sp_0_eq}
            \centering
            \boldsymbol{x}_{0,{{k}{|}{k}}} = \hat{\boldsymbol{x}}_{{k}{|}{k}}
        \end{equation}
         \begin{equation}\label{sp_i_eq}
            \centering
            \begin{aligned}
            \boldsymbol{x}_{i,{{k}{|}{k}}} =& \hat{\boldsymbol{x}}_{{k}{|}{k}} + (\sqrt{({n}+\lambda)\boldsymbol{\mathbf{P}}_{{k}{|}{k}}})_{i},
            \\& {{i}= {1},...,{n}}
            \end{aligned}
       \end{equation}
        \begin{equation}\label{sp_n_i_eq}
            \begin{aligned}
            \centering
            \boldsymbol{x}_{i,{{k}{|}{k}}} =& \hat{{x}}_{{k}{|}{k}} - (\sqrt{({n}+\lambda)\boldsymbol{\mathbf{P}}_{{k}{|}{k}}})_{i-n}, \\
            & {{i}= {n+1},...,{2n}}
            \end{aligned}
        \end{equation}
        where $\boldsymbol{\mathbf{P}}_{{k}{|}{k}}$ is positive definite, therefore it can be factored by Cholesky decomposition as $\sqrt{\boldsymbol{\mathbf{P}}_{{k}{|}{k}}}\sqrt{\boldsymbol{\mathbf{P}}_{{k}{|}{k}}}^{T}$.
        $\sqrt{\boldsymbol{\mathbf{P}}_{{k}{|}{k}}}_i$ is the ${i}$ row of the matrix $\sqrt{\boldsymbol{\mathbf{P}}_{{k}{|}{k}}}$.
        The points are symmetrical around the expected value, so the expectation is preserved.
        ${\lambda=\alpha^2(n+\kappa)-n}$ is a scaling parameter, $\alpha$ determines the spread of the sigma points around $\hat{{x}}_{{k}{|}{k}}$, usually set to a small value ($1e-3$), $\kappa$ is a secondary scaling parameter usually set to 0, and $\beta$ is used to incorporate prior knowledge of the distribution of $x$ (for Gaussian distribution it is set to 2).        
    \item Time update\\
        The sigma points (vectors) propagate through the dynamic model mapping, after which we compute the reflected mean and covariance as follows:
        \begin{equation}\label{tu_x_dynam_eq}
            \centering
            \boldsymbol{x}_{i,{{k+1}{|}{k}}} = f(\boldsymbol{x}_{i,{{k+1}{|}{k}}}), i=0,...,2n
        \end{equation}
        \begin{equation}\label{tu_x_mean_eq}
            \centering
            \hat{\boldsymbol{x}}_{{k+1}{|}{k}} = \sum_{i=0}^{2n} \mathbf{w}_{i}^{m} \boldsymbol{x}_{i,{{k+1}{|}{k}}}
        \end{equation}
        \begin{equation}\label{tu_p_eq}
            \centering
            \boldsymbol{\mathbf{P}}_{{k+1}{|}{k}}=
            \sum_{i=0}^{2n}\mathbf{w}_{i}^{c}{\delta\boldsymbol{x}_{i,{{k+1}{|}{k}}} \cdot {\delta\boldsymbol{x}_{i,{{k+1}{|}{k}}}}^T} + \boldsymbol{\mathbf{Q}}_{k+1}
        \end{equation}
        where $(\circ)^T$ is the transform operator, $\delta\boldsymbol{x}_{i,{{k+1}{|}{k}}} = \boldsymbol{x}_{i,{{k+1}{|}{k}}}-\hat{\boldsymbol{x}}_{{k+1}{|}{k}}$, $\boldsymbol{\mathbf{Q}}_{k+1}$ is the covariance process noise matrix at step ${k+1}$

    \item Measurement update\\
        Update the sigma points according to the time update phase to estimate the observation/measurement:
        \begin{equation}\label{sp_0_meas_eq}
            \centering
            \boldsymbol{x}_{0,{{k+1}{|}{k}}} = \hat{{x}}_{{k+1}{|}{k}}
        \end{equation}
        \begin{equation}\label{sp_i_meas_eq}
            \centering
            \begin{aligned}
            \boldsymbol{x}_{i,{{k+1}{|}{k}}} =& \hat{\boldsymbol{x}}_{{k+1}{|}{k}} + (\sqrt{({n}+\lambda)\boldsymbol{\mathbf{P}}_{{k+1}{|}{k}}})_{i},
            \\& {{i}= {1},...,{n}}
            \end{aligned}
        \end{equation}
        \begin{equation}\label{sp_n_i_meas_eq}
            \centering
            \begin{aligned}
            \boldsymbol{x}_{i,{{k+1}{|}{k}}} =& \hat{\boldsymbol{x}}_{{k+1}{|}{k}} - (\sqrt{({n}+\lambda)\boldsymbol{\mathbf{P}}_{{k+1}{|}{k}}})_{i-n},\\& {{i}= {n+1},...,{2n}}
            \end{aligned}
        \end{equation}
        Compute the estimated measurement of each sigma point:
        \begin{equation}\label{z_i_eq}
            \centering
            \boldsymbol{z}_{i,{{k+1}{|}{k}}} = h(\boldsymbol{x}_{i,{{k+1}{|}{k}}}), i=0,...,2n 
        \end{equation}
        Compute the mean estimated measurement:
        \begin{equation}\label{z_mean_eq}
            \centering
            \hat{\boldsymbol{z}}_{{k+1}{|}{k}} = \sum_{i=0}^{2n}{w}_{i}^{m}\boldsymbol{z}_{i,{{k+1}{|}{k}}}
        \end{equation}
        Compute the measurement covariance matrix:
        \begin{equation}\label{s_eq}
            \centering
            \boldsymbol{\mathbf{S}}_{k+1} = \sum_{i=0}^{2n}{w}_{i}^{c}\delta\boldsymbol{z}_{i,{{k+1}{|}{k}}}\cdot {\delta\boldsymbol{z}_{i,{{k+1}{|}{k}}}}^T+\boldsymbol{\mathbf{R}}_{k+1}
        \end{equation}
        where $\delta\boldsymbol{z}_{i,{{k+1}{|}{k}}} = \boldsymbol{z}_{i,{{k+1}{|}{k}}} - \hat{\boldsymbol{z}}_{{k+1}{|}{k}}$, $\boldsymbol{\mathbf{R}}_{k+1}$ is the measurement process noise at step ${k+1}$.
        
        Compute the cross-covariance matrix:
        \begin{equation}\label{p_cross_cov_eq}
            \centering
            \boldsymbol{\mathbf{P}}_{k+1}^{x,z} = \sum_{i=0}^{2n}{w}_{i}^{c}\delta\boldsymbol{x}_{i,{{k+1}{|} {k}}}\cdot \delta\boldsymbol{z}_{i,{{k+1}{|}{k}}}^T
        \end{equation}
        where $\delta\boldsymbol{x}_{i,{{k+1}{|}{k}}} = \boldsymbol{x}_{i,{{k+1}{|}{k}}}-\hat{\boldsymbol{x}}_{{k+1}{|}{k}}$ and $\delta\boldsymbol{z}_{i,{{k+1}{|}{k}}} = \boldsymbol{z}_{i,{{k+1}{|}{k}}} - \hat{\boldsymbol{z}}_{{k+1}{|}{k}}$.\\
        Compute the Kalman gain:
        \begin{equation}\label{k_eq}
            \centering
            \boldsymbol{\mathbf{K}}_{k+1} = \boldsymbol{\mathbf{P}}_{k+1}^{x,z}\boldsymbol{\mathbf{S}}_{k+1}^{-1}
        \end{equation}    
        Compute the new mean according to the observation $z_{k+1}$:
        \begin{equation}\label{x_meas_eq}
            \centering
            \hat{\boldsymbol{x}}_{{k+1}{|}{k+1}}=\hat{\boldsymbol{x}}_{{k}{|}{k+1}} + \boldsymbol{\mathbf{K}}_{k+1}(z_{k+1}-\hat{\boldsymbol{z}}_{{k+1}{|}{k}})
        \end{equation}    
        Compute the new covariance matrix using the Kalman gain:
        \begin{equation}\label{P_mean_eq}
            \centering
            \boldsymbol{\mathbf{P}}_{{k+1}{|}{k+1}} = \boldsymbol{\mathbf{P}}_{{k+1}{|}{k}}-\boldsymbol{\mathbf{K}}_{k+1}\boldsymbol{\mathbf{S}}_{k+1}{\boldsymbol{\mathbf{K}}_{k+1}}^{T} 
        \end{equation}
\end{enumerate}
    Equation \eqref{tu_x_dynam_eq} is the sigma points propagating equation where $f(\boldsymbol{x}_{i,{{k+1}{|}{k}}})$ is the dynamic model mapping. Using accurate dynamic mapping is crucial for the accuracy and appropriate behavior of the filter. The proposed method introduces an accurate implementation of the dynamic model for a navigation error state vector.

\subsection{The Inertial Navigation Algorithm}\label{sd_alg}
In each cycle, the inertial navigation algorithm integrates the inertial sensors outputs to produce the integrated solution. It includes the position vector, given in geodetic form (latitude, longitude, altitude) [rad, rad, m], the velocity vector [m/s] in north, east, down (NED) coordinate system, and the transformation from the body frame to the navigation frame.
The inertial navigation algorithm can be found in many books, as in \cite{groves2008}. The formulation for the transition from time step $k$ to time step $k+1$ is as follows:
\begin{itemize}
    \item \textbf{Update the attitude} Given ${C_{b,k}^n}$ - the transformation matrix from the body frame to the navigation frame at time step $k$ and $w_{ib,k+1}^b$ - the angular velocity sensed by the Gyroscopes during the last time step, given in the body frame, calculate the intermediate rotation from the body frame to the navigation frame $C_{b,k+1}^n$ as follows:
    \begin{equation}\label{sd_att_updt}
    {C^*}_{b,k+1}^n = C_{b,k}^n e^{\check{{w^*}_{nb,k+1}^b} \frac{dt}{2}}
    \end{equation}
    where $\check{}$ is the skew-symmetric operator, $dt$ is the cycle duration ($\frac{dt}{2}$ was used to calculate the average attitude during step $k+1$ to project the acceleration measurements from body frame to the navigation frame), and 
    \begin{equation}\label{sd_att_w_nb}
    {w^*}_{nb,k+1}^b = w_{ib,k+1}^b - C_{n,k}^b (w_{ie,k}^n + w_{en,k}^n)
    \end{equation}
    where $w_{ib,k+1}^b$ is the angular velocity of the body frame relate to the inertial system, resolved in the body frame, during time step $k+1$ in the body frame, given by the gyroscopes, $w_{ie,k}^n$ is the earth rotation relate to the inertial frame, given in the navigation frame computed at time step $k$, and $w_{en,k}^n$ is the angular velocity of the navigation frame relate to earth resolved in the navigation frame computed at time step $k$. The symbol $*$ denotes intermediate calculation phase.
    \item \textbf{Update the velocity}:
    \begin{equation}\label{sd_v_dot}
    \centering
    \dot{v}_{k+1}^n = {C^*}_{b,k+1}^n{f_{ib,k+1}^b}+g^n-(\check{w_{en,k}^n}+ 2\check{w_{ie,k}^n})v_k^n
    \end{equation}
    where $f_{ib}^b$ is the accelerometers measured specific force in the body frame respect to the inertial frame resolved in the body frame, $\check{w_{en}^n}$ and $\check{w_{ie}^n}$ are the skew-symmetric matrices of the transport rate and the earth rotation correspondingly, resolved in the navigation frame, computed at time step $k+1$ based on time step $k$, $v_k^n$ is the velocity of the body relate to earth at time step $k$. The updated velocity is: 
    \begin{equation}\label{sd_v}
    v_{k+1}^n = v_k^n + \dot{v}_{k+1}^n dt
    \end{equation}
    Note that $v^n = [{v_n^n}, {v_e^n}, {v_d^n}]^T$ the velocity in the north, east and down axes.
    \item Update the position:
    \begin{equation}\label{sd_pos_h}
    \centering
    h_{k+1} = h_{k} - ({{v_d^n}}_{k+1} + {{v_d^n}}_k)\frac{dt}{2}
    \end{equation}
    \begin{equation}\label{sd_pos_lat}
    L_{k+1} = L_{k} + \left( 
    \frac{{{v_n^n}}_k}{R_n + h_k} + \frac{{{v_n^n}}_{k+1}}{R_n + h_{k+1}} \right)\frac{dt}{2}
    \end{equation}
    \begin{equation}\label{sd_pos_long}
    \lambda_{k+1} = \lambda_{k} + \left(
    \frac{{{v_e^n}}_k}{R_e + h_k} + \frac{{v_e^n}_{k+1}}{R_e + h_{k+1}} \right)
    \frac{dt}{2}
    \end{equation}
    where $h$ is altitude, $L$ is the latitude, $\lambda$ is the longitude, $R_n$ and $R_e$ are the meridian and transverse radii of curvature correspondingly.
    \item \textbf{Update the attitude}:
    \begin{equation}\label{sd_att_updt_final}
    C_{b,k+1}^n = C_{b,k}^n  e^{\check{w_{nb,k+1}^b}  dt}
    \end{equation}
    and $w_{nb,k+1}^b$ is computed using the most updated position and velocities computed in \eqref{sd_v} - \eqref{sd_pos_long}:
    \begin{equation}\label{sd_att_w_nb_final}
    w_{nb,k+1}^b= w_{ib,k+1}^b - C_{n,k}^b (w_{ie,k+1}^n + w_{en,k+1}^n)
    \end{equation}
\end{itemize}      
We shall call the above procedure the navigation algorithm cycle and we shall use it to propagate the UKF sigma points, during the time update step.

\section{Proposed Method}\label{prop_appr}
The propagation of the UKF sigma points when applying \eqref{tu_x_dynam_eq} is crucial for the precision of the calculations of the propagated mean and covariance matrix. Each sigma point represents an error state vector:
\begin{equation}\label{err_state_eq}
    \centering
    \delta{\boldsymbol{x}}=\begin{bmatrix}
    \delta{\boldsymbol{v}^n} & \boldsymbol\delta{\Psi}^n & \boldsymbol{b}_a & \boldsymbol{b}_g
    \end{bmatrix}^T \in \mathbb{R}^{12 \times 1}
\end{equation}
where $\delta{\boldsymbol{v}^n}\in\mathbb{R}^3$ is the velocity error state vector, expressed in the navigation frame, $\boldsymbol\delta\Psi^n \in \mathbb{R}^3$ is the misalignment error state vector, expressed in the navigation frame, $\boldsymbol{b}_a\in\mathbb{R}^3$ is the accelerometers residual bias vector expressed in the body frame, and $\boldsymbol{b}_g\in\mathbb{R}^3$ is the gyroscope's residual bias vector expressed in the body frame. Having 12 states requires 25 sigma points, each holding 12 elements. There are two common ways to propagate the sigma points: 1) a linearized dynamic model as described in \cite{groves2008} and 2) A non-linearized dynamic model \cite{AUVHypersonic2020}. Both approaches try to imitate the propagation of the errors states when applying the navigation algorithm. Our proposed approach offers to propagate the sigma points by directly using the strapdown navigation algorithm, including intermediate updated computations, to capture the most accurate system behavior. To this end, we introduce the following method to propagate sigma points at time step $k+1$.
\begin{enumerate}
\item At time step $k+1$ each sigma point, is added to the mean solution at time step $k$, yielding a solution that represents the specific  sigma point.
\item The modified solution is propagated via the navigation algorithm cycle(s), using the inertial measurements. This results with 25 modified solutions, propagated to time step $k+1$. The first one is the propagation of the mean solution. To extract the propagated sigma points, we calculate the difference between the propagated mean solution and each of the modified propagated solutions. Since the biases remain constant they are not propagated. Yet, they are used to compensate the inertial measurements that the navigation algorithm uses when propagating the solution of each sigma point.
\end{enumerate}
To define the above formally, let ${x}_k$ be the solution of the navigation algorithm at time step $k$:
\begin{equation}\label{x_mean_eq}
    \centering
    {x}_k=\begin{bmatrix}
    \xi & v^n & \Psi^n
    \end{bmatrix}^T \in \mathbb{R}^{9 \times 1}
\end{equation}
where $\xi \in \mathbb{R}^{3 \times 1}$ is the position vector consisting of the latitude, longitude, and altitude, $v^n\in \mathbb{R}^{3 \times 1}$ is the velocity expressed in the navigation frame, and, $\Psi^n\in \mathbb{R}^{3 \times 1}$ is the Euler angles representing the transformation from the navigation frame to the body frame.
Let ${\sigma}_k\in \mathbb{R}^{12 \times 1}$ be a given sigma point (error state vector) at time step $k$ such that:
\begin{equation}\label{x_sp_i_pos_eq}
    \centering
    {{x_{\sigma}}^\xi}_k = {{x}^\xi}_k
\end{equation}
\begin{equation}\label{x_sp_vel_eq}
    \centering
    {{x_{\sigma}}^v}_k = {{x}^v}_k + {{\sigma}^{\delta v}}_k
\end{equation}
\begin{equation}\label{x_sp_i_psi_eq}
    \centering
    {{x_{\sigma}}^\psi}_k = {{x}^\psi}_k \oplus {{\sigma}^{\delta\psi}}_k
\end{equation}
where $\oplus$ denotes addition via transformation matrix and ${x_{\sigma}}_k$ is a representative solution of the sigma point $\sigma$. We shall compute ${x_{\sigma}}_{k+1}$ using the navigation algorithm cycle (\eqref{sd_att_updt} - \eqref{sd_att_w_nb_final}) and label it $na\_cycle$. To apply the navigation algorithm cycle, from step $k$ to step $k+1$, the accelerations and gyroscopes measurements $f_{ib}^b, w_{ib}^b$ (For brevity, we omit the subscript $ib$ in the rest of the derivation) are required. Since the sigma points hold also the estimated bias errors, we need to compensate the measurements with the appropriate biases before applying the navigation cycle, reflecting the sigma point states. Thus using \eqref{x_sp_i_pos_eq} - \eqref{x_sp_i_psi_eq} we can write:
\begin{equation}\label{x_modified_eq}
    \centering
    {x_{\sigma}}_{k+1} = na\_cycle({x_{\sigma}}_{k}, {{f^*}^b}_{k+1}, {{w^*}^b}_{k+1})
\end{equation}
where ${{f^*}^b}_{k+1}$ and ${{w^*}^b}_{k+1}$ are the inertial measurements compensated by the appropriate sigma point biases.
To extract the propagated sigma point, we need to subtract from each solution the propagated mean solution, denoted as $x_{m}$, resulting with the propagated sigma points:
\begin{equation}\label{sp_dv_extract_eq}
    \centering
    {\sigma}^{\delta v}_{k+1} = {x_{\sigma}^v}_{k+1} - {x_{m}^v}_{k+1} 
\end{equation}
\begin{equation}\label{sp_dpsi_extract_eq}
    \centering
    {\sigma}^{\delta \psi}_{k+1} = {x_{\sigma}^\psi}_{k+1} \ominus {x_{m}^\psi}_{k+1}
\end{equation}
where $\ominus$ denotes subtraction via transformation.\\
Once the sigma points \eqref{sp_dv_extract_eq}-\eqref{sp_dpsi_extract_eq} are obtained, they are used in \eqref{tu_x_mean_eq}-\eqref{p_cross_cov_eq} with the standard UKF implementation.

\section{Experimental Results}\label{res_sec}
\subsection{Dataset}\label{exp_res_dataset}
\begin{figure}[h]
    \centering
    \centering
    \includegraphics[width=5cm, height=4cm]{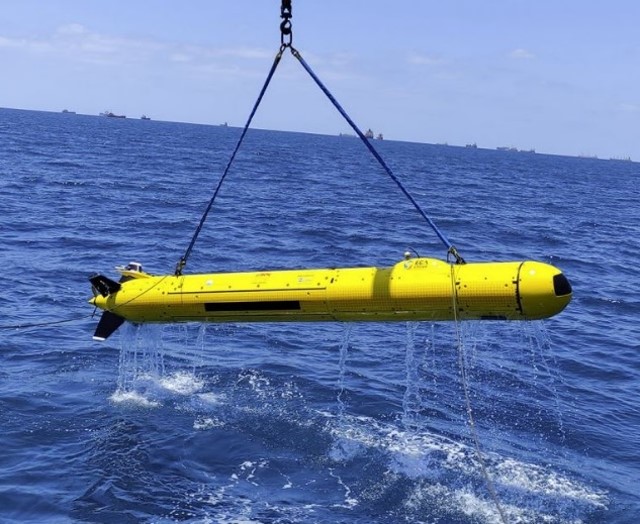}
    \caption{The Snapir AUV during sea experiments.}
    \label{fig:Snapir-AUV}
\end{figure}

We used the Snapir AUV (Figure \ref{fig:Snapir-AUV}) dataset \cite{10674766} for training and testing our proposed approach. Snapir is an ECA Robotics modified Group A18D mid-size AUV. The Snapir AUV is outfitted with the iXblue, Phins Subsea INS, which uses fiber optic gyroscope for precise inertial navigation \cite{iXblue}. Snapir also uses a Teledyne RDI Work Horse Navigator Doppler velocity log (DVL) \cite{TeledyneMarine}, known for its capability to provide accurate velocity measurements. The INS operates at a frequency of 100 [Hz], whereas the DVL operates at 1[Hz].

\begin{figure}[h]
    \centering
    \begin{subfigure}[b]{0.45\linewidth}
        \centering
        \includegraphics[width=4.3cm, height=2.0cm]{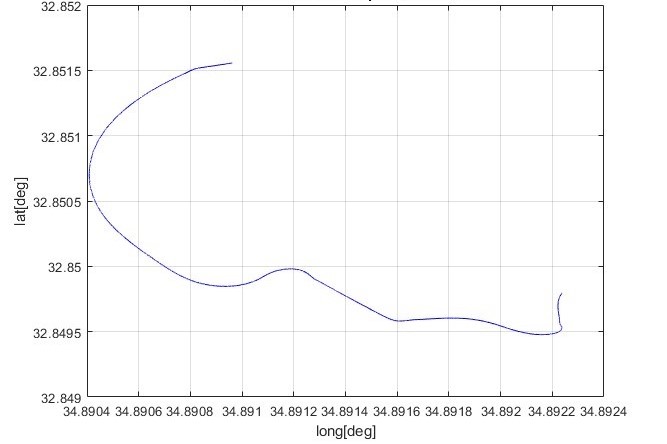}
        \caption{Track 1}
        \label{fig:track1}
    \end{subfigure}
    \hfill
    \begin{subfigure}[b]{0.45\linewidth}
        \centering
        \includegraphics[width=4.3cm, height=2.0cm]{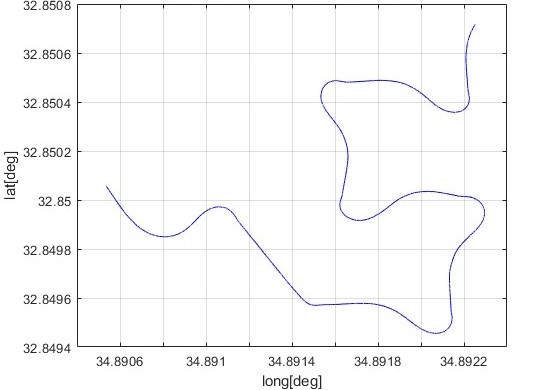}
        \caption{Track 2}
        \label{fig:track2}
    \end{subfigure}
    \hfill
    \caption{Horizontal position of tracks 1-2 used in the training dataset.}
    \label{fig:horiz-pos-tracks-1-2}
\end{figure}

\begin{figure}[h]
    \begin{subfigure}[b]{0.45\linewidth}
        \centering
        \includegraphics[width=4.3cm, height=2.0cm]{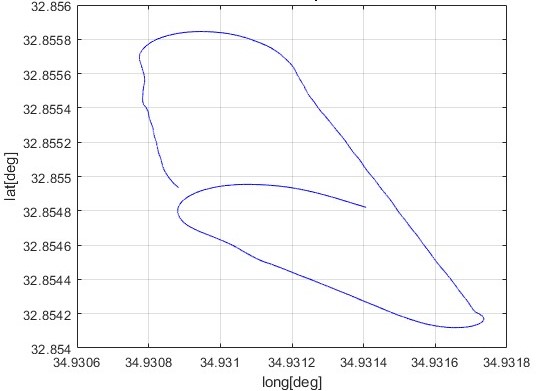}
        \caption{Track 3}
        \label{fig:track3}
    \end{subfigure}
    \hfill
    \begin{subfigure}[b]{0.45\linewidth}
        \centering
        \includegraphics[width=4.3cm, height=2.0cm]{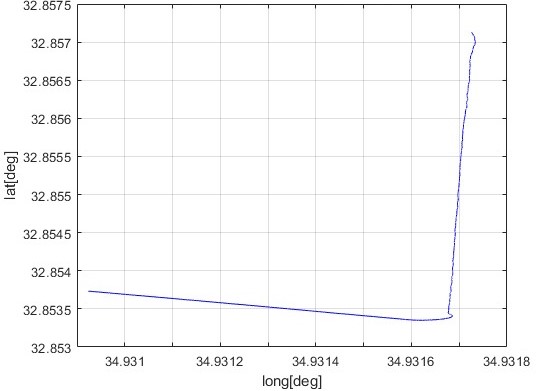}
        \caption{Track 4}
        \label{fig:track4}
    \end{subfigure}
    \caption{Horizontal position of tracks 3-4 used in the training dataset.}
    \label{fig:horiz-pos-tracks-3-4}
\end{figure}

The dataset contains 24 minutes of recording divided into 6 trajectories. Each trajectory contains  the DVL/INS fusion solution which is the ground truth (GT), inertial readings, and DVL measurements. To create the trajectories being tested, we added the following to the given raw data: initial velocity errors with a standard deviation (STD) of $0.25[m/s]$ and $0.05[m/s]$, horizontal and vertical, respectively, and misalignment errors with a STD of $0.01[deg]$ per axis. We also added noises with a STD of $0.03[m/s^2]$ to the accelerometers and $7.3\times 10^{-6}[rad/s]$ to the gyroscopes. Finally, we added biases with a STD of $0.3[m/s^2]$ to the accelerometers and $7.3\times 10^{-5}[rad/s]$ to the gyroscopes.\\
For the training process, we used tracks 1-4, each consisting of 4 minutes, resulting in a total of 16 minutes of the training dataset. The trajectories presented in Figure \ref{fig:horiz-pos-tracks-1-2} and Figure \ref{fig:horiz-pos-tracks-3-4} include a wide range of dynamics and maneuvers.
Track 1 (Figure \ref{fig:track1}) and track 2 (Figure \ref{fig:track2}) have relatively more maneuvers and dynamic changes than track 3 (Figure \ref{fig:track3}) and track 4 (Figure \ref{fig:track4}), has longer straight sections.

\begin{figure}[h]
    \centering
    \begin{subfigure}[b]{0.45\linewidth}
        \centering
        \includegraphics[width=4.3cm, height=2.0cm]{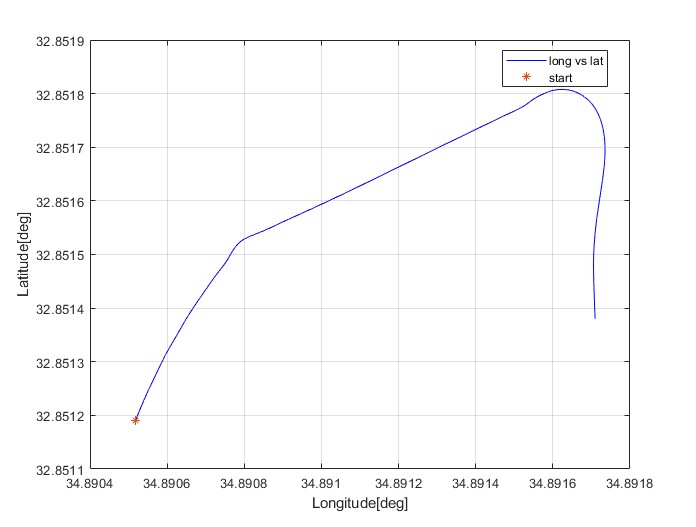}
        \caption{Track 5}
        \label{fig:track5}
    \end{subfigure}
    \hfill
    \begin{subfigure}[b]{0.45\linewidth}
        \centering
        \includegraphics[width=4.3cm, height=2.0cm]{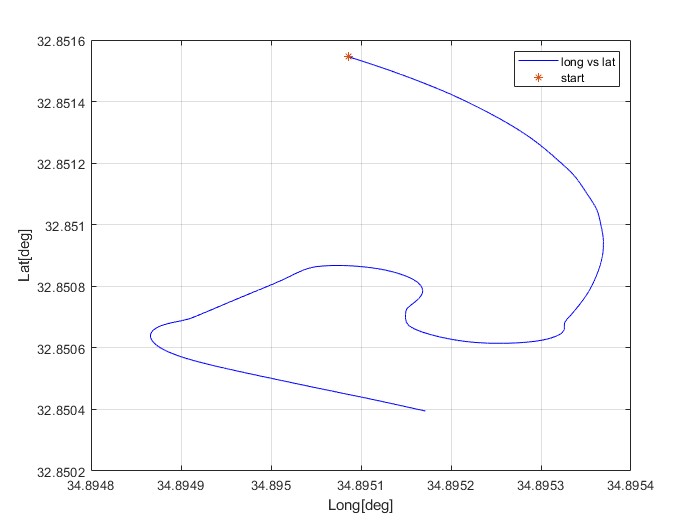}
        \caption{Track 6}
        \label{fig:track6}
    \end{subfigure}
    \caption{Horizontal position of tracks 5-6 used in the testing dataset.}
    \label{fig:horiz-pos-tracks-5-6}
\end{figure}

The testing datasets included tracks 5-6, presented in Figure \ref{fig:horiz-pos-tracks-5-6}, with a total time of 3.5 minutes.
\subsection{Evaluation metrics and baseline}\label{exp_res_matrices}
In \cite{ANUKF2025}, the authors showed that adaptive neural UKF (ANUKF) with dynamic process noise performs better than the adaptive EKF that uses neural network and from the standard non-adaptive UKF. Therefore, we used ANUKF as the baseline and compared it with an ANUKF that uses the proposed UKF-NESPM, denoted as ANUKF-NESPM.\\
For a fair comparison, the filters used the same navigation algorithm. The navigation algorithm initialized by the reference solution integrates the updated IMU outputs and reduces navigation errors by fusing the DVL measurements. Moreover, the filters had the same training process as our proposed network architecture.\\
For each track we calculated the total velocity root mean square error (VRMSE) and the total misalignment root mean square error (MRMSE).
For a given track, the VRMSE is calculated as follows:
\begin{equation}\label{vrmse_calc}
    \centering
    VRMSE = \sqrt{\frac{1}{n}\sum_{i=1}^{n}\|{\delta \boldsymbol{v}(i)}\|^2}
\end{equation}
where, $n$ is the number of samples in the trajectory, $\|{\delta \boldsymbol{v} (i)}\|$ is the norm of the velocity error between the estimated velocity and the GT in time step $i$. The track average is:
\begin{equation}\label{vrmse_avg_calc}
    \centering
    VRMSE_{avg} = \sqrt{\frac{1}{2}({VRMSE_{5}}^2 + {VRMSE_{6}}^2)}
\end{equation}
where $VRMSE_{5}$ and $VRMSE_{6}$ are the $VRMSE$ of tracks 5 and 6, respectively.\\
In the same manner, the MRMSE is:
\begin{equation}\label{Mrmse_calc}
    \centering
    MRMSE = \sqrt{\frac{1}{n}\sum_{i=1}^{n}\|{\delta {\boldsymbol{\Psi}}(i)}\|^2}
\end{equation}
where ${\delta {\boldsymbol{\Psi}}(i)}$ is the misalignment vector that represents by the Euler angles corresponding to the following transformation matrix $\boldsymbol{\mathbf{C}}_{n_r}^n$:
\begin{equation}\label{misaligment_err_calc}
    \centering
    \boldsymbol{\mathbf{C}}_{n_r}^n = \boldsymbol{\mathbf{C}}_b^n \cdot \boldsymbol{\mathbf{C}}_{n_r}^b
\end{equation}
and $\boldsymbol{\mathbf{C}}_{n_r}^b$ is the GT rotation matrix from the navigation frame to the body frame.
\subsection{Results}\label{exp_res_res}
We applied the two filters, ANUKF and ANUKF-NESPM, to the test dataset. Table \ref{tab:dvl_full_comp_tbl} summarizes the VRMSE and MRMSE results of the two filters, showing that the velocity solution of the ANUKF-NESPM performs better than the ANUKF. In other words, ANUKF-NESPM showed an improvement of $8.8\%$ over ANUKF. The ANUKF-NESPM estimation of misalignment showed an improvement of $17.7\%$ over ANUKF.
\begin{table}[h]
    \caption{VRMSE and MRMSE of the ANUKF, and ANUKF-NESPM applied to the testing dataset.}
    \centering
    \begin{tabular}{|p{2.3cm}|p{1cm}|p{1cm}|p{1cm}|p{1cm}|}
    \hline
    Method & VRMSE [m/sec] & baseline improv. & MRMSE [rad] & baseline improv. \\
    \hline
    \textbf{ANUKF-NESPM(ours)} & 0.073 &   - & 0.0065 &   - \\
    \textbf{ANUKF(baseline)} & 0.08 &   8.8\% & 0.0079 &   17.7\% \\
    \hline
    \end{tabular}
    \label{tab:dvl_full_comp_tbl}
\end{table}
%
\section{Conclusion}\label{conc_sec}
Accurate estimation of the propagated mean and covariance matrix in the presence of nonlinear dynamic model is crucial for the correct and accurate behavior of the UKF. To cope with that, we introduced a new method that utilizes the navigation algorithm to propagate the sigma points, improving the accuracy of the propagated estimations.\\
To demonstrate the performance of our approach, we focused on the INS/DVL fusion problem for a maneuvering AUV. We evaluated our approach compared to the adaptive neural UKF using a real-world recorded AUV dataset. In the VRMSE metric, ANUKF-NESPM improved by $8.8\%$ over ANUKF. In the MRMSE metric, ANUKF-NESPM improved by $17.7\%$ over ANUKF.
In conclusion our ANUKF-NESPM offers an accurate navigation solution in normal operating conditions of INS/DVL fusion. Thus, it allows planning and operating in challenging AUV tasks, including in varying dynamics.
%
\section*{Acknowledgements}
The authors gratefully acknowledge the support of the Israel Innovation Authority under grant 84330, which partially supported this research.

\bibliographystyle{IEEEtran}
\bibliography{pIBio}

\end{document}